\documentclass[9pt]{article}
\usepackage{spconf,amsmath,graphicx}
\usepackage{subfigure}
\usepackage{setspace}
\usepackage{amssymb}
\usepackage{graphicx,psfrag} 
\usepackage{epsfig} 
\usepackage{esdiff} 
\usepackage{amsmath,amssymb,mathabx} 
\usepackage{xfrac}   
\usepackage{pst-sigsys} 
\usepackage{algorithm}
\usepackage{cite}
\usepackage{lipsum}
\usepackage{algorithmic}
\usepackage{colortbl}
\usepackage{float}
\usepackage{setspace}
\usepackage{color}

\newcommand{\beq}{\begin{equation}}
\newcommand{\eeq}{\end{equation}}
\newcommand{\beqa}{\begin{eqnarray}}
\newcommand{\eeqa}{\end{eqnarray}}
\newcommand{\ben}{\begin{enumerate}}
\newcommand{\een}{\end{enumerate}}

\newcommand\blfootnote[1]{%
  \begingroup
  \renewcommand\thefootnote{}\footnote{#1}%
  \addtocounter{footnote}{-1}%
  \endgroup
}

\title{Online Reweighted Least Squares Algorithm for Sparse Recovery and Application to Short-Wave Infrared Imaging}

\namelineone{Subhadip Mukherjee$^{\star}$ \qquad Deepak R.$^{\star}$ \qquad Huaijin Chen$^{\dagger}$ \qquad Ashok Veeraraghavan$^{\dagger}$}
\namelinetwo{Chandra Sekhar Seelamantula$^{\star}$}
\address{$^{\star}$Department of Electrical Engineering, Indian Institute of Science, Bangalore, Karnataka, India\\$^{\dagger}$Department of Electrical and Computer Engineering, Rice University, Houston, Texas, USA\\Emails: \{subhadip, drm94, chandra.sekhar\}@ee.iisc.ernet.in, \{huaijin.chen, vashok\}@rice.edu}

\begin{document}
%
\maketitle
%


\small
\begin{abstract}
We address the problem of sparse recovery in an online setting, where random linear measurements of a sparse signal are revealed sequentially and the objective is to recover the underlying signal. We propose a reweighted least squares (RLS) algorithm to solve the problem of online sparse reconstruction, wherein a system of linear equations is solved using conjugate gradient with the arrival of every new measurement. The proposed online algorithm is useful in a setting where one seeks to design a progressive decoding strategy to reconstruct a sparse signal from linear measurements, so that one does not have to wait until all measurements are acquired. Moreover, the proposed algorithm is also useful in applications where it is infeasible to process all the measurements using a batch algorithm, owing to computational and storage constraints. It is not needed a priori to collect a fixed number of measurements; rather one can keep collecting measurements until the quality of reconstruction is satisfactory and stop taking further measurements once the reconstruction is sufficiently accurate. We provide a proof-of-concept by comparing the performance of our algorithm with the RLS-based batch reconstruction strategy, known as \textit{iteratively reweighted least squares} (IRLS), on natural images. Experiments on a recently proposed focal plane array-based imaging setup shows up to $1$ dB improvement in output peak signal-to-noise ratio as compared with the total variation-based reconstruction.\blfootnote{\scriptsize\hspace{-0.2in}This work is partly supported by the DRDO-IISc FRONTIERS research program.}
\end{abstract}

\section{Introduction}
Compressive sensing (CS) \cite{candesintro2008,baraniuk1,donoho_it}, a signal processing paradigm that combines the acquisition and compression of signals by exploiting sparsity, has attracted enormous attention in the signal processing community over the past decade. The key objective of CS is to recover a signal $\boldsymbol x \in \mathbb{R}^n$ from an incomplete noisy linear projection $\boldsymbol y=\boldsymbol A \boldsymbol x+\boldsymbol \xi \in \mathbb{R}^m$, $m<n$, by imposing the constraint of sparsity.~Seminal work by Cand\`es et al. \cite{candes_romberg_tao} has shown that the seemingly challenging combinatorially hard problem of signal recovery can be solved by a tractable convex program under mild restrictions, namely the \textit{restricted isometry property} (RIP), on the sensing matrix $\boldsymbol A$. Two major algorithmic paradigms have been developed for CS: (i) greedy algorithms such as orthogonal matching pursuit \cite{tropp_omp}, CoSaMP \cite{tropp_cosamp} etc. where the signal support is estimated in a greedy manner, followed by the estimation of amplitudes using least-squares and (ii) convex relaxation-based approaches such as LASSO \cite{tibshirani}, where a sparsity promoting convex regularization is incorporated to solve the otherwise ill-posed inverse problem.\\
\indent The conventional CS deals with signal reconstruction in batch, wherein the entire set of measurements is available at one's disposal. However, in most real applications, the measurements are acquired sequentially and it is important to develop online algorithms that reconstruct the underlying signal gradually as the measurements arrive. Moreover, in an online setting, the number of measurements does not have to be fixed in advance and one can stop as and when the quality of reconstruction is acceptable.~Initial contribution to sequential CS was made by Malioutov et al. \cite{malioutov}, who showed that a stopping rule can be designed for collecting measurements guaranteeing accurate reconstruction. Asif and Romberg \cite{asif_romberg} developed dynamic algorithms based on homotopy continuation for sequential CS assuming that the underlying signal changes slowly during measurement acquisition. Homotopy-based algorithms for solving online LASSO have also been developed in \cite{homotopy1,homotopy2}. Angelosante et al. proposed time- and norm-weighted LASSO schemes \cite{twl} where the $\ell_1$-norm weights are obtained from the recursive least-squares algorithm. Sequential reconstruction of sparse signals with slowly varying sparsity patterns has also been addressed by Vaswani et al. \cite{vaswani_kfcs} and has been applied to dynamic magnetic resonance imaging. Their technique goes by name of Kalman-filter CS (KFCS). 

The developments in the theory of CS have led to remarkable advances in imaging technologies in recent years \cite{willett,romberg_imaging}. CS asserts that using the sparsity prior, one can reconstruct a high-resolution (HR) image from its coded low-resolution (LR) measurements, thereby effectively increasing the sampling rate. Duarte et al. made pioneering contribution in compressed imaging by developing \textit{single pixel cameras} (SPCs) \cite{spc}, wherein one acquires coded measurements of a HR scene using a single photo detector and subsequently recovers the image by employing a CS reconstruction algorithm. Imaging in the short-wave infrared (SWIR) spectrum offers several benefits over visible imaging and is used in many applications. SWIR penetrates fog and smog, enables passive imaging in the dark and has a variety of applications in biomedical imaging \cite{fpa_cs_rice}. Cameras for SWIR, however, are much more expensive than their counterparts in the visible spectrum. Motivated by the Single Pixel Camera (SPC) \cite{spc}, Chen \textit{et al}.  \cite{fpa_cs_rice} developed a proof-of-concept prototype hardware for compressive imaging in the SWIR spectrum. The setup, which is termed the \emph{Focal Plane Array-based Compressive Sensing} (FPA-CS), captures several low-resolution measurements with its $64\times64$ SWIR sensor array, which are used for reconstructing an image of much higher resolution. We show that the proposed online approach fits naturally in the FPA-CS imaging framework and demonstrate results on a simulated setup.\\
\indent Our main contribution is to propose an online algorithm for sparse reconstruction by exploiting the idea of \textit{iteratively reweighted least squares} (IRLS), originally proposed by Daubechies et al. \cite{irls_daubechies}. We refer to the proposed algorithm as online reweighted least squares (ORLS). Akin to IRLS, the key idea in ORLS is to approximate the sparsity inducing $\ell_1$ norm with a weighted $\ell_2$ penalty, where the weights are determined based on the current estimate of the signal. The underlying signal is assumed to be fixed during the course of measurement. With each new measurement, a system of linear equations is solved using the conjugate-gradient (CG) method, which requires less number of iterations to converge as more measurements are made available. We demonstrate that the ORLS algorithm is noise-robust and produces reconstructed signals on par with the batch-mode recovery using IRLS. Experiments on FPA-CS image reconstruction show that ORLS leads to an improvement of approximately $1$ dB in peak signal-to-noise ratio (PSNR) over the total variation (TV) regularization-based batch recovery. 
  
\section{Problem Statement and the Proposed ORLS Algorithm}
Our objective is to recover a sparse signal $\boldsymbol x^* \in \mathbb{R}^n$ from its noisy linear projections of the form $y_t=\boldsymbol  a_t^\top \boldsymbol  x^*+\xi_t$, revealed sequentially at time instants $t=1,2,\cdots$. Under this setting, we formulate the problem of sequential sparse recovery as an online convex optimization problem, where one predicts a vector $\boldsymbol x_t$ in a convex set $S$, and suffers a convex loss $f_t\left(\boldsymbol x_t\right)$, revealed at every time instant $t$. Naturally, the loss function is defined as $f_t\left(\boldsymbol x\right)=\left(y_t-\boldsymbol a_t^\top \boldsymbol x\right)^2$, so that it measures how well the current estimate explains the measurements. The set $S$ is defined as $S\stackrel{\Delta}{=}\left\{\boldsymbol x \in \mathbb{R}^n: \left\|\boldsymbol x \right\|_1\leq \tau  \right\}$, the closed $\ell_1$-norm ball of radius $\tau$, to promote sparsity in the estimate. To perform online sparse estimation, it would be natural to solve
\begin{equation}
\boldsymbol x_{t+1} = \arg \underset{\boldsymbol x \in S}{\min}\sum_{j=1}^{t}\left(y_j-\boldsymbol a_j^\top\boldsymbol x\right)^2,
\label{FTL_for_online_CS}
\end{equation}
where $\boldsymbol x_{t+1}$ denotes the reconstructed sparse vector at time instant $t$. Solving (\ref{FTL_for_online_CS}) yields estimates $\boldsymbol x_t$ that are consistent with the measurements received until time $t$, and are within the set $S$, thereby promoting sparsity. 
\subsection{Online reweighted least-squares (ORLS) algorithm}
In order to develop the ORLS algorithm, we write (\ref{FTL_for_online_CS}) in its equivalent unconstrained form, given by 
\begin{equation}
\boldsymbol x_{t+1} = \arg \underset{\boldsymbol x \in \mathbb{R}^n}{\min}\sum_{j=1}^{t}\left(y_j-\boldsymbol a_j^\top\boldsymbol x\right)^2 +\lambda \left\|\boldsymbol x\right\|_1,
\label{ORLS_step1}
\end{equation}
where $\lambda>0$ is an appropriately chosen regularization parameter which trades-off between sparsity and data fidelity. The problem in (\ref{ORLS_step1}) is a convex program. However, because of the $\ell_1$ regularizer, it is not possible to find a closed-form solution to (\ref{ORLS_step1}). To circumvent this problem, we propose to approximate the $\ell_1$ norm using a quadratic of the form $\boldsymbol x^\top\boldsymbol W_{t} \boldsymbol x$, where $\boldsymbol W_t$ is a diagonal matrix, containing positive weights on its diagonal. The entries of $\boldsymbol W_t$ are suitably updated based on the previous estimate $\boldsymbol x_t$, using the formula $\boldsymbol W_{t}\left(j\right)\leftarrow \displaystyle \frac{1}{\left| \boldsymbol x_{t}\left(j\right) \right|+\delta}$, for every $j$, where $\delta>0$ is a small positive constant that is introduced to avoid numerical instability. Therefore, at every $t$, one needs to solve the following optimization problem: 
\begin{equation}
\boldsymbol x_{t+1} = \arg \underset{\boldsymbol x \in \mathbb{R}^n}{\min}\sum_{j=1}^{t}\left(y_j-\boldsymbol a_j^\top \boldsymbol x\right)^2 +\lambda \boldsymbol x^\top \boldsymbol W_t \boldsymbol x.
\label{ORLS_step2}
\end{equation}
The closed-form solution to (\ref{ORLS_step2}) is given by
\begin{equation}
\boldsymbol x_{t+1}=\left(\lambda \boldsymbol W_t + \sum_{j=1}^{t}\boldsymbol a_j \boldsymbol a_j^\top\right)^{-1}\left( \sum_{j=1}^{t}y_j \boldsymbol a_j\right).
\label{ORLS_step3}
\end{equation}
The matrix inversion in (\ref{ORLS_step3}) can be performed using the Sherman-Morrison\footnote{Sherman-Morrison formula: $\left(\boldsymbol A+ \boldsymbol u \boldsymbol v^\top\right)^{-1}=\boldsymbol A^{-1}-\frac{\boldsymbol A^{-1}\boldsymbol u \boldsymbol v^\top \boldsymbol A^{-1}}{1+\boldsymbol v^\top \boldsymbol A^{-1} \boldsymbol u}$} formula recursively, requiring $\mathcal{O}\left(n^2t\right)$ computations, provided that $\boldsymbol a_i$, $i=1:t$, are stored. Thus, for $m$ measurements, the overall run-time complexity of the ORLS algorithm is given by $T_{\text{ORLS}}\left(n\right)=\sum_{t=1}^{m}\mathcal{O}\left(n^2t\right)=\mathcal{O}\left(n^2 m^2\right)$. If $\boldsymbol a_i$s are not stored individually and one only keeps track of the accumulated value $\sum_{j=1}^{t}\boldsymbol a_j \boldsymbol a_j^\top$, the inversion in (\ref{ORLS_step3}) demands $\mathcal{O}\left(n^3\right)$ complexity, which is unacceptably high for an online scheme. To circumvent this problem, we propose to find the estimate $\boldsymbol x_{t+1}$ using the CG method. Computing $\boldsymbol x_{t+1}$ using (\ref{ORLS_step3}) is equivalent to solving the system of linear equations given by   
\begin{equation}
\boldsymbol A_t \boldsymbol x =\boldsymbol b_t,
\label{ORLS_CG}
\end{equation}
where $\boldsymbol A_t$ and $\boldsymbol b_t$, for $t\geq 2$, are computed recursively:
\begin{equation}
\boldsymbol A_t = \lambda\boldsymbol W_t + \boldsymbol Q_{t-1}+\boldsymbol a_t \boldsymbol a_t^\top \text{\,\,and\,\,}\boldsymbol b_t = \boldsymbol b_{t-1} +y_t\boldsymbol a_t, 
\label{Ab_update}
\end{equation}
by setting $\boldsymbol Q_{1}=\boldsymbol a_1 \boldsymbol a_1^\top$ and $\boldsymbol b_{1} =y_1 \boldsymbol a_1$. To evaluate $\boldsymbol x_{t+1}$ using CG, we set the previous estimate $\boldsymbol x_{t}$ as the initialization. Each iteration of CG requires $\mathcal{O}\left(n^2\right)$ computation and one does not need to store the previous values of $y_i$ and $\boldsymbol a_i$. An overall saving in computation and storage is realized if the number of iterations $L$ required in the CG algorithm is smaller than $m$. Typically, we observe that $L=\frac{n}{2}$ suffices to obtain equivalent performance as the ORLS algorithm.  

\begin{figure}[t]
\centering
\begin{tabular}{ccccc}
\includegraphics[width=1.5in]{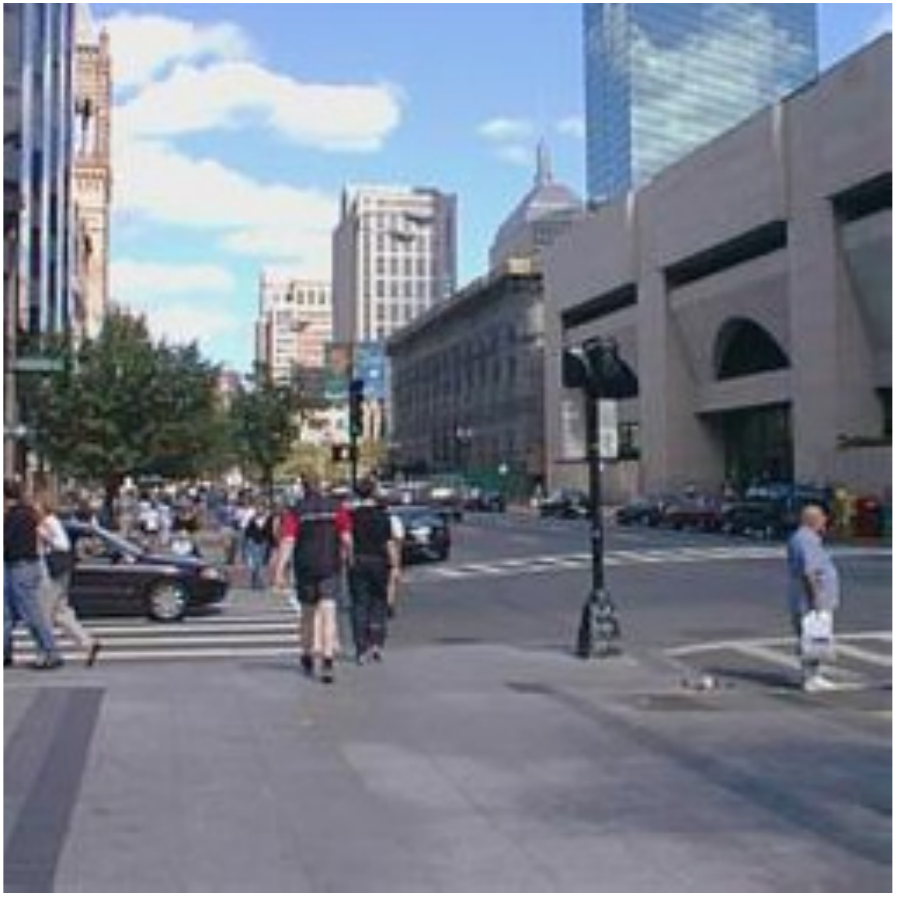}&
\includegraphics[width=1.5in]{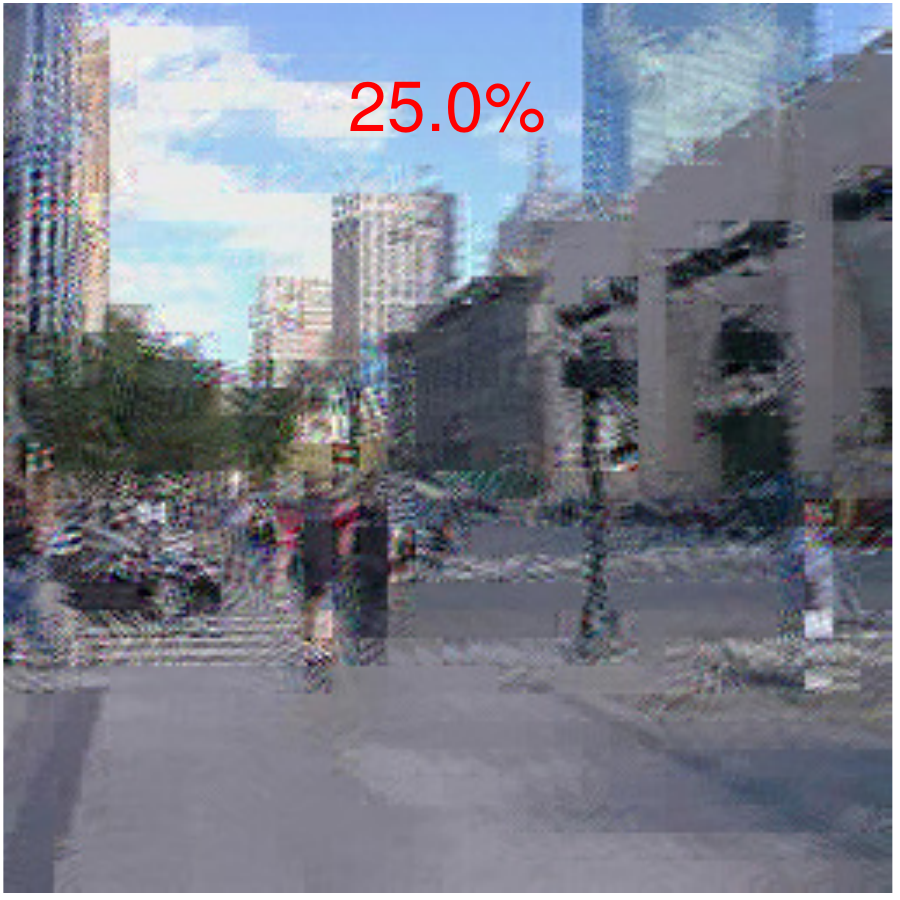}\\
\small {(a)} &\small{(b)}\\
\includegraphics[width=1.5in]{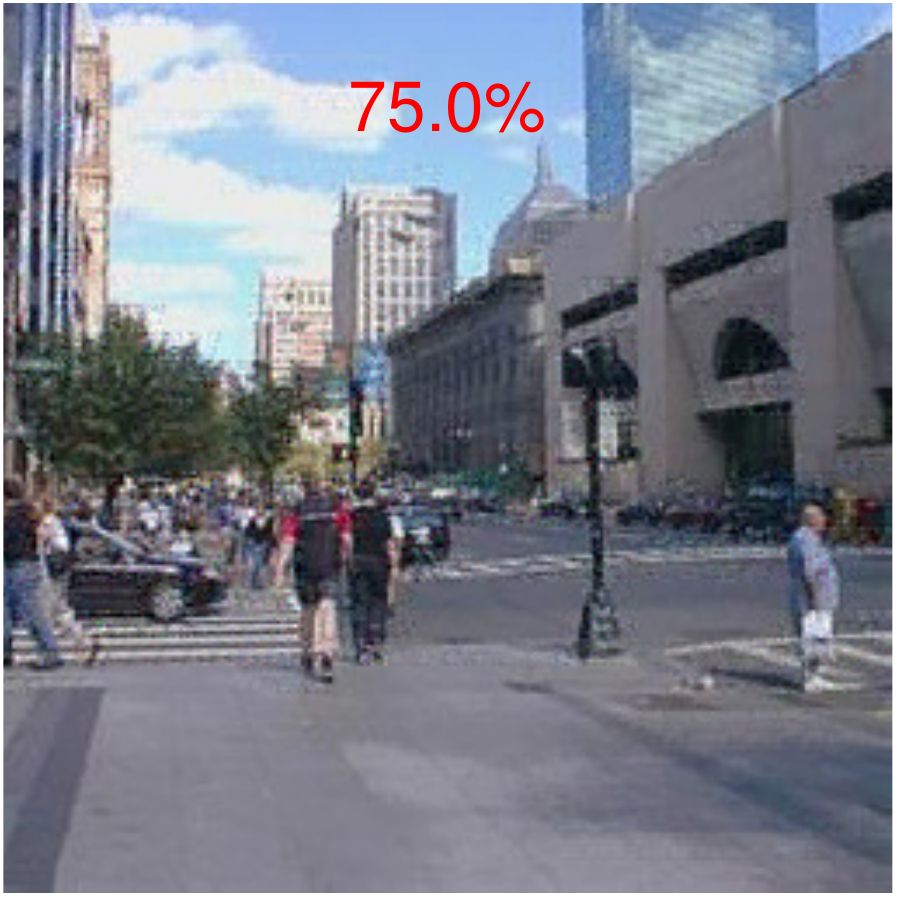}&
\includegraphics[width=1.5in]{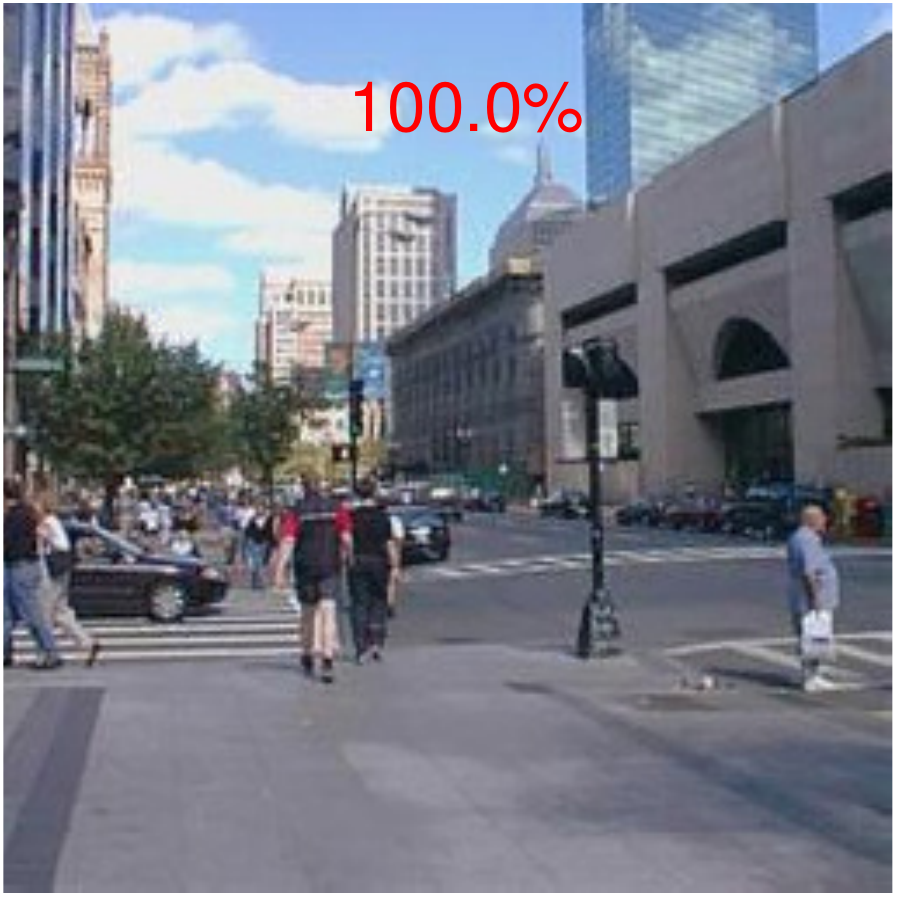}\\
\small {(c)} & \small{(d)}\\
\includegraphics[width=1.5in]{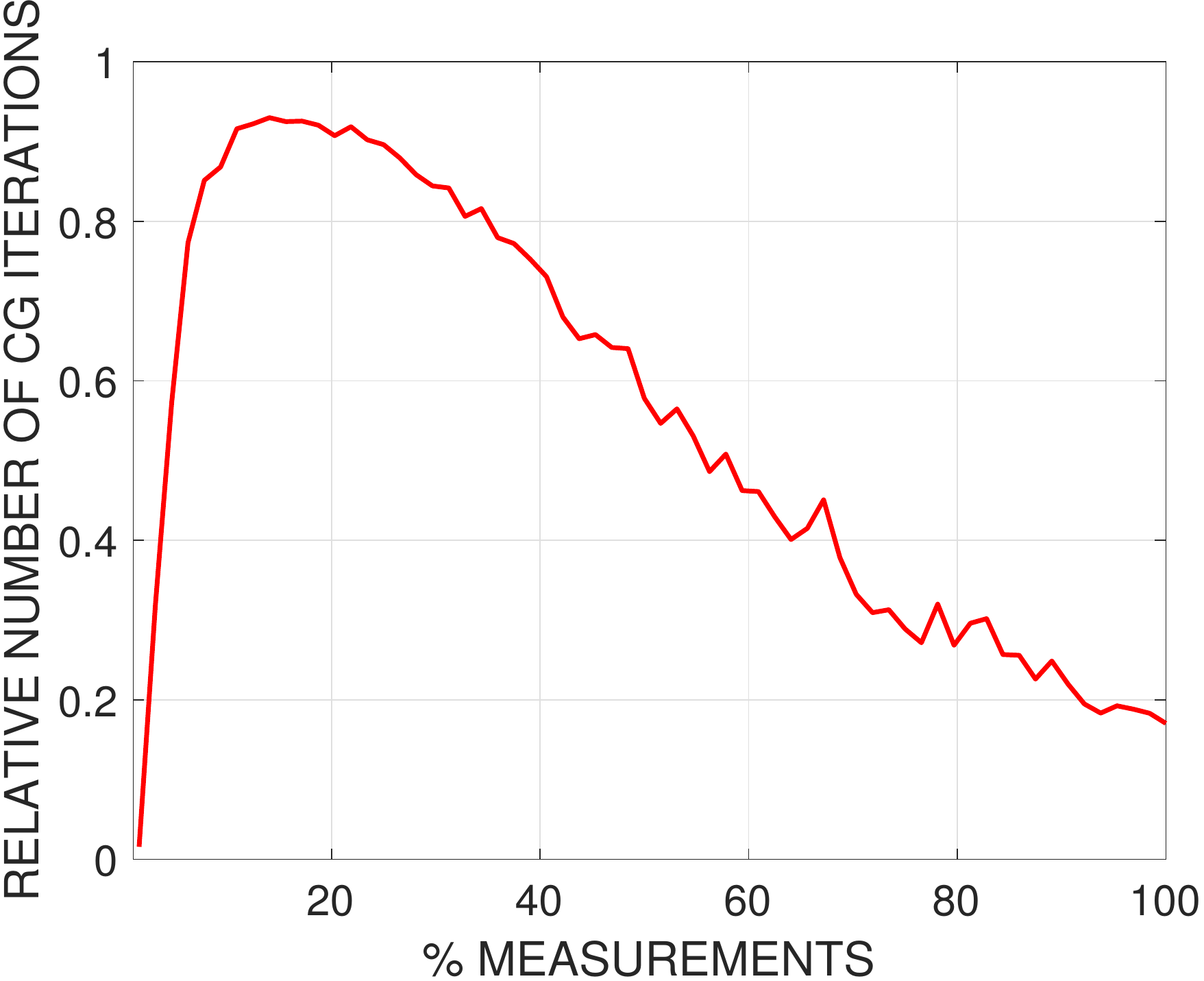}&
\includegraphics[width=1.5in]{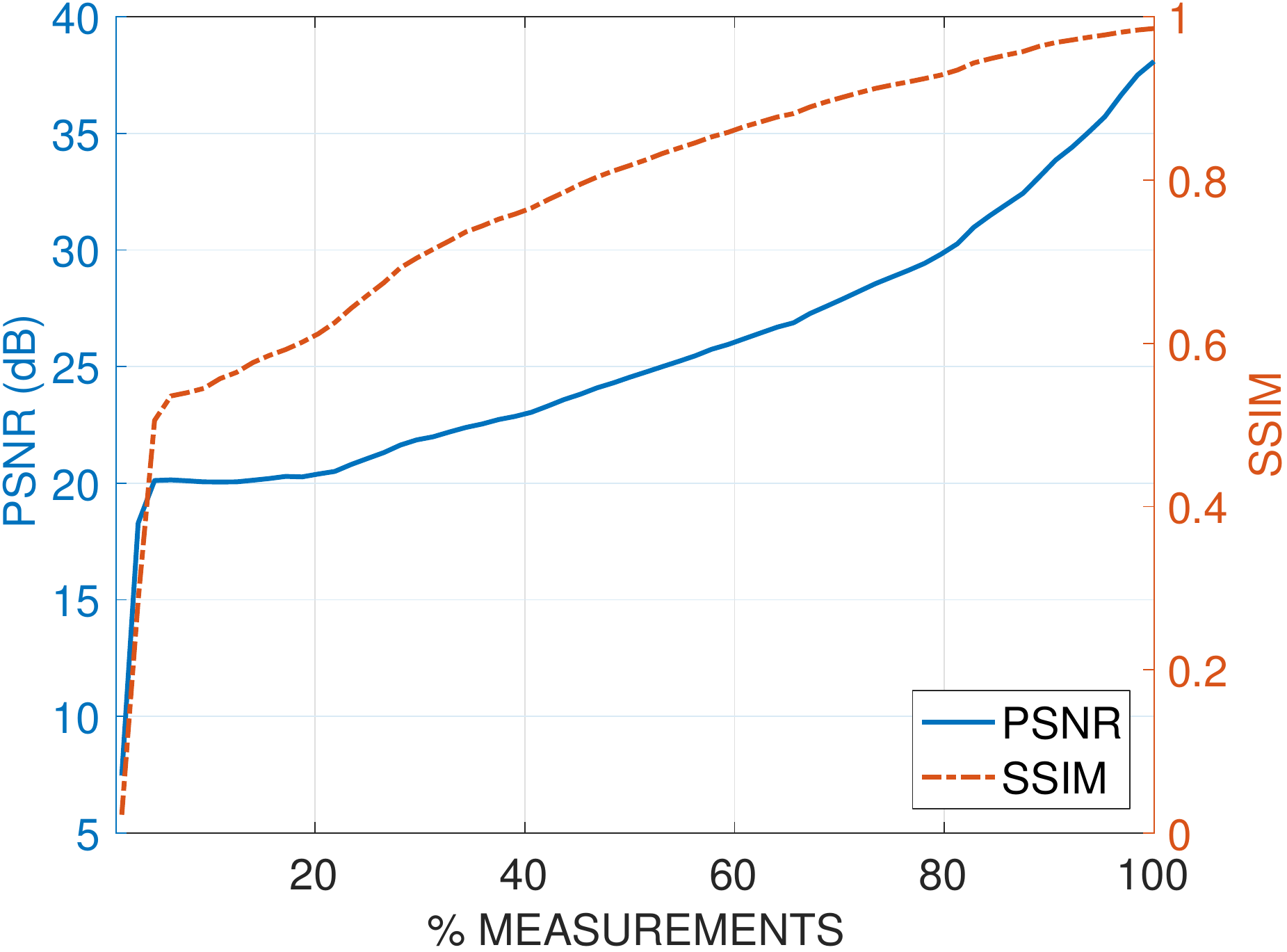}\\
 \small{(e)} &\small{(f)}\\
\end{tabular}
\caption{\small{(Color online) Reconstruction of images using ORLS: (a) corresponds to the ground-truth; (b), (c), and (d) correspond to the reconstructed images with $25$\%, $75$\%, and $100$\% measurements, respectively. The number of CG iterations required for convergence is shown in (e), whereas the evolutions of PSNR and SSIM with respect to measurements are plotted in (f).}}
\label{ORLS_noiseless_images_fig}
\end{figure}
\begin{figure}[t]
\centering
\subfigure[Batch IRLS]{
\includegraphics[width=1.5in]{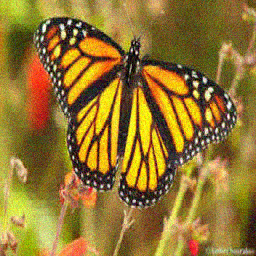}
}
\subfigure[ORLS]{
\includegraphics[width=1.5in]{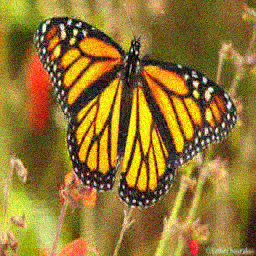}
}\\
\subfigure[Absolute difference between (a) and (b)]{
	\includegraphics[width=1.25in]{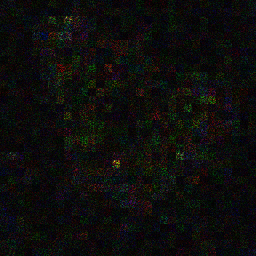}
}
\subfigure[PSNR and SSIM versus \% measurements]{
\includegraphics[width=1.75in]{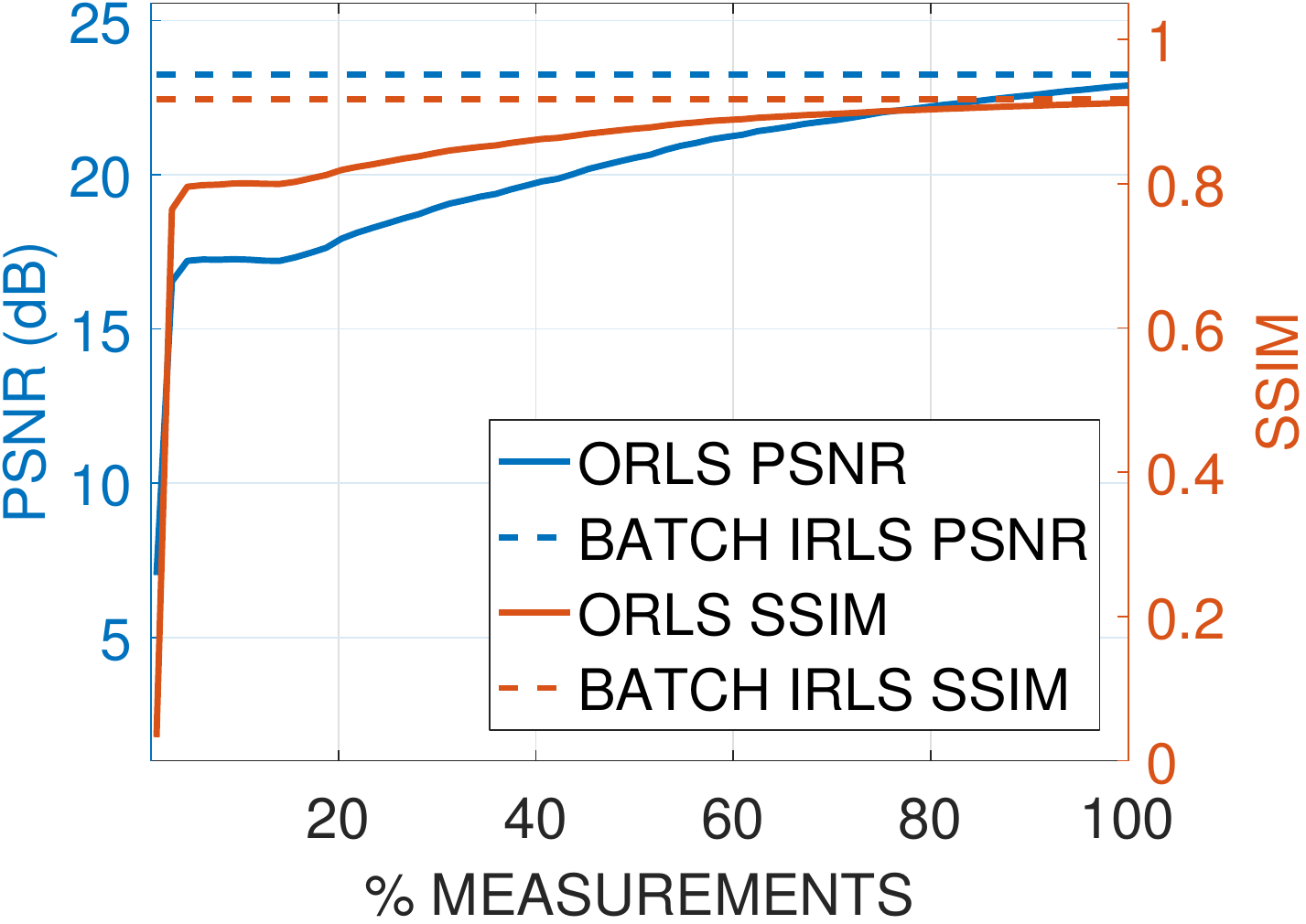}
}\\
\caption{\small{(Color online) Comparison of ORLS with batch IRLS. The PSNR and SSIM of ORLS are on par with their batch counterparts.}}
\label{online_versus_batch_fig}
\end{figure}

\section{Experimental Results}
To image a scene, random binary patterns of size $8\times 8$ consisting of zeros and ones are used for scanning. Such patterns can be produced in practice by employing programmable digital micro-mirror devices (DMDs), wherein one can program the binary patterns and acquire spatially coded measurements. Using random patterns of zeros and ones amounts to acquiring the summation of intensities of a random subset of pixels over every $8\times 8$ patch of the HR scene. We denote a generic HR patch by $\boldsymbol z$, and assume that it admits a sparse representation of the form $\boldsymbol z=\boldsymbol D\boldsymbol x$, where $\boldsymbol D$ is the DCT dictionary. The ORLS algorithm recovers $\boldsymbol z$ from sequentially obtained coded measurements of the form $y_t=\boldsymbol c_t^\top\boldsymbol z=\boldsymbol a_t^\top\boldsymbol x$, where $\boldsymbol a_t = \boldsymbol D^\top \boldsymbol c_t$ and $\boldsymbol c_t$ denotes the random binary pattern. Each patch contains $n=64$ pixels and $\left\{\frac{t}{n}\right\}_{t=1}^{m}$ denotes the fraction of measurements acquired until time $t$. Experiments are conducted on color images and the reconstruction performance is shown for $m$ varying from $1$ to $n$. Performance metrics for evaluating the quality of reconstruction are taken as the PSNR and the structural similarity index (SSIM), considering the ground-truth image as the reference.\\
\indent The reconstructed images using ORLS corresponding to $25$\%, $75$\%, and $100$\% measurements are shown in Figures~\ref{ORLS_noiseless_images_fig}(b), \ref{ORLS_noiseless_images_fig}(c), and \ref{ORLS_noiseless_images_fig}(d), respectively. The ground-truth image is shown in Fig. \ref{ORLS_noiseless_images_fig}(a) to facilitate visual comparison. We consider noiseless coded measurements in this experiment and the parameter $\lambda$ is set to $1$. The reconstruction in Fig. \ref{ORLS_noiseless_images_fig}(d) corresponding to the full set of measurements is indistinguishable from the ground-truth with no color artifacts. This is also reflected in the PSNR and SSIM values plotted as a function of percentage measurements in Fig. \ref{ORLS_noiseless_images_fig}(f). The values of PSNR and SSIM corresponding to $100$\% measurements are approximately $38$ dB and $0.99$, indicating a near-accurate recovery. We have also shown the number of iterations required by the CG algorithm (normalized by $n$) in order to achieve an accuracy of $\epsilon=10^{-5}$ with respect to the fraction of measurements. It is observed that the number of iterations required for convergence increases initially and attains a peak, thereafter falling almost monotonically as more measurements arrive. This trend indicates that a considerable saving in computation can be achieved over the direct inversion-based approach by employing CG to solve \eqref{ORLS_CG}.\\
\indent We compare the performance of ORLS with its batch counterpart, namely IRLS, with the full set of measurements ($m=64$ per patch).~The HR scene to be recovered is corrupted with additive Gaussian noise with a PSNR of $22.10$ dB. The reconstructed images obtained using the batch IRLS and ORLS algorithms corresponding to $\lambda=40$ are shown in Figures~\ref{online_versus_batch_fig}(a) and \ref{online_versus_batch_fig}(b), respectively. The difference image is shown in Figure \ref{online_versus_batch_fig}(c). The values lie in the range $[0, 76]$ and are scaled in the figure for better visualization. The variations of PSNR and SSIM of the output of ORLS are plotted in Fig. \ref{online_versus_batch_fig}(d) as a function of measurements. We observe that ORLS leads to a reconstruction with PSNR and SSIM matching that of batch IRLS (shown using dotted horizontal lines), indicating that ORLS performs on par with its batch counterpart.         
\begin{figure}[t]
\centering
\begin{tabular}{ccccc}
\includegraphics[width=3.0in]{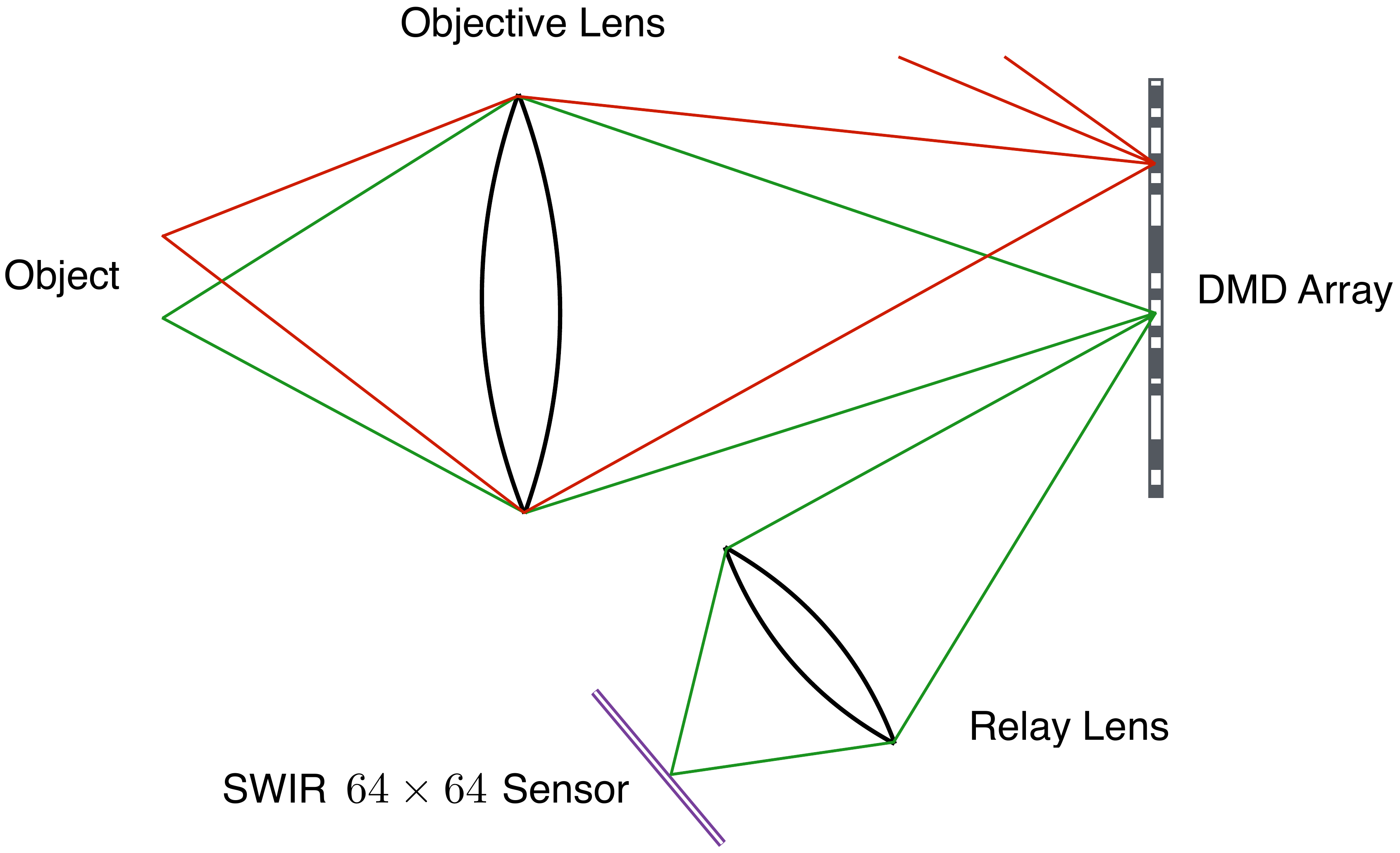}
\end{tabular}
\caption{\small{A schematic of the FPA-CS imaging setup.}}
\label{fpa_cs_setup_fig}
\end{figure}

\section{Application of ORLS to FPA-CS}
Focal-plane-array-based CS (FPA-CS) is the architecture proposed in \cite{fpa_cs_rice} that leverages CS for short-wave infrared (SWIR) imaging. It is optically identical to several SPCs arranged in an array and provides increased measurement rates, thereby resulting in higher spatio-temporal resolution.~A schematic of the FPA-CS imaging setup in shown in Fig.~\ref{fpa_cs_setup_fig}. The object to be imaged is focused onto a DMD array, which acts as a random binary mask on the image. The light from the DMD is refocused by the relay lens onto a $64\times 64$ SWIR sensor array. Several such LR images of the scene are obtained on the SWIR sensor array for a given object, each with a different binary mask. The set of masks is also stored and is used in the reconstruction along with the LR measurements.\\
\indent The key objective is to recover the image formed on the DMD array, which we shall refer to as the \textit{input image}. Since the output measurements are downsampled versions of the input image, every output sensor captures light from a small patch of the DMD. Thus every output pixel is obtained from a small patch of the input image. Let $\boldsymbol{z}$ be the input image of size $M\times N$ formed on the DMD plane and $\boldsymbol{y}_t$ be the $t^{\text{th}}$ observed image of size $m\times n$ formed on the SWIR sensor array, vectorized so that $\boldsymbol{y}_t\in \mathbb{R}^{mn}$. Each pixel in $\boldsymbol{y}_t$ corresponds to a set of pixels or a patch of the image $\boldsymbol z$. We denote the $p^{\text{th}}$ pixel on $\boldsymbol{y}_t$ as $y_t(p)$, $p = 1,\cdots,mn$, and assume that it corresponds to the $P^{\text{th}}$ patch on $\boldsymbol{z}$, denoted as $\boldsymbol{z}(P)$. The intensity measured by the output pixel $y_t(p)$ is an accumulated effect of the binary pattern applied by the DMDs and blurring on $\boldsymbol{z}(P)$:  
\begin{equation*}
y_t(p)=\boldsymbol{c}(P)^\top \boldsymbol{z}(P)+{\xi}_t(P),
\end{equation*}
where $\boldsymbol{c}(P)$ represents the combined effect of blurring and the DMD pattern. The binary mask can in fact be chosen to be periodic without loss of generality, so that it is same for every patch $P$, thus requiring less storage. Substituting $\boldsymbol{z}(P)=\boldsymbol D \boldsymbol{x}(P)$, where $\boldsymbol D$ is the DCT dictionary and $\boldsymbol {x}(P)$ is sparse, we have that 
\begin{equation}
y_t(p)=\boldsymbol{a}(P)^\top \boldsymbol{x}(P)+{\xi}_t(P), t=1,2,\cdots
\label{eqn:orls-fpacs-io}
\end{equation}
where $\boldsymbol a(P)= \boldsymbol D^\top \boldsymbol c (P)$. The modified measurement equation \eqref{eqn:orls-fpacs-io} reveals that the image reconstruction task in FPA-CS fits naturally in the framework considered in ORLS. Notably, the patches $\boldsymbol{z}(P)$ of the input image might have overlaps and one should compute the average of the neighboring reconstructed patches on overlapping regions to form the input image. For our experiment, we have simulated the FPA-CS imaging setup such that there is no overlap of the patches in $\boldsymbol z$. Since the acquisition for each patch is independent of the others, it is possible to parallelize the patch-wise computations.

\begin{figure}[t]
	\centering
	\begin{tabular}{ccccc}
		\includegraphics[width=1.5in]{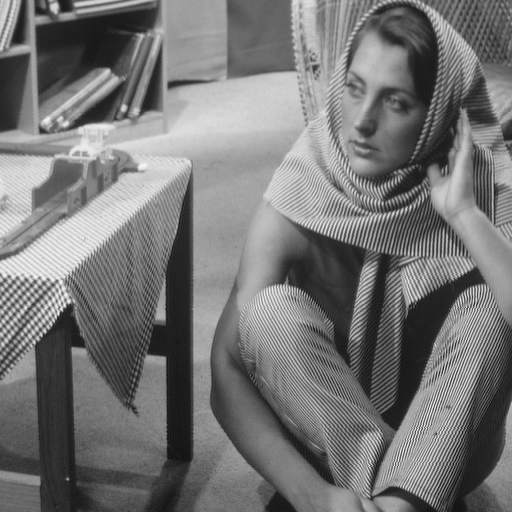}&
		\includegraphics[width=1.5in]{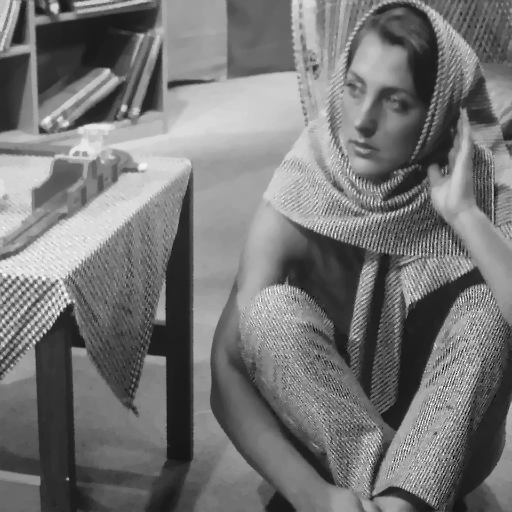}\\
		\small {(a) Ground-truth} &\small{(b) TV reconstruction in batch}\\
		\includegraphics[width=1.5in]{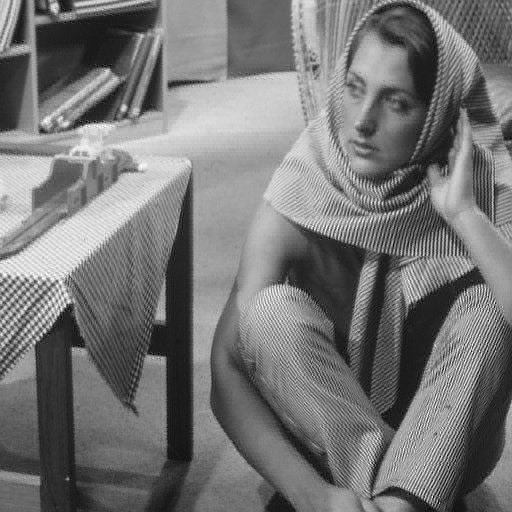}&
		\includegraphics[width=1.5in]{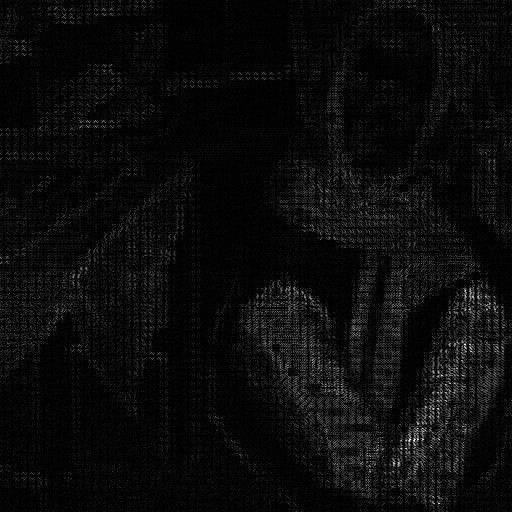}\\
		\small {(c) ORLS} & \small{(d) Difference between (a) and (c)}\\
	\end{tabular}
	\caption{\small{Comparison of ORLS with batch-mode reconstruction based on TV regularization for FPA-CS.}}
	\label{fig:FPA-CS}
\end{figure}

Experiments are conducted on a simulated FPA-CS setup and the performance of ORLS is compared with the TV regularization-based batch mode reconstruction as in \cite{fpa_cs_rice}. The patch size on the input image is taken as $8\times8$. The results are reported in Fig.~\ref{fig:FPA-CS}, which shows a comparison of the ORLS reconstruction (Fig.~\ref{fig:FPA-CS}(c)) using 64 measurements (which corresponds to $100$\% of the measurements), with the TV-based batch reconstruction (Fig.~\ref{fig:FPA-CS}(b)). We observe that ORLS leads to a reconstruction with better visual quality, which is reflected in higher PSNR ($28.99$ dB versus $27.98$ dB) and SSIM ($0.89$ versus $0.88$) values. The difference between the ground-truth and the ORLS-based reconstruction is shown in Fig.~\ref{fig:FPA-CS}(d) with intensity values scaled to fall in the range $[0,113]$ to ensure better visual assessment.  We observe that although some textures in the target image are not recovered accurately, most of the edge information is reliably reconstructed. 

\section{Conclusions and Outlook}
We have developed an online regularized least squares algorithm for sparse reconstruction. The ORLS algorithm solves a system of linear equation using CG with the arrival of each new measurement. We have demonstrated that the CG algorithm takes considerably less number of iterations than the signal dimension if the current estimate of the signal is used as the initialization. This leads to progressively less amount of computation as more measurements are acquired. We did not consider measurement adaptation in our formulation and assumed that the signal remains fixed during the measurement process. Numerical experiments on natural images show that the ORLS scheme is robust to noise and reconstructs images that are on par with the batch IRLS technique in terms of PSNR and SSIM. The ORLS reconstruction is also free from any color artifacts. We have demonstrated that an improvement of $1$ dB in output PSNR can be achieved using ORLS as compared with the batch-mode reconstruction that uses the TV regularizer.

\bibliographystyle{IEEEbib}
\bibliography{strings,refs}

\vfill
\pagebreak

\bibliographystyle{IEEEbib}
\bibliography{strings,refs}

\end{document}